\def\year{2022}\relax
\documentclass[letterpaper]{article} 
\usepackage{aaai22}  
\usepackage{times}  
\usepackage{helvet}  
\usepackage{courier}  
\usepackage[hyphens]{url}  
\usepackage{graphicx} 
\usepackage{natbib}  
\usepackage{caption} 
\DeclareCaptionStyle{ruled}{labelfont=normalfont,labelsep=colon,strut=off} 
\usepackage{lipsum}
\usepackage{amsmath} 
\usepackage{paralist} 
\usepackage{booktabs}
\usepackage{rotating}
\usepackage{array}
\usepackage[capitalise,nameinlink]{cleveref}
\usepackage{amsthm}
\newtheorem{definition}{Definition}

\newcommand\pic[2][1]{\includegraphics[width=#1\linewidth,trim=2cm 2cm 2cm 2cm,clip]{fig/toy_sample_images/#2}}
\usepackage{tikz}
\newcommand{\igwc}[5]{\begin{tikzpicture}
	\draw (0, 0) node[inner sep=0] {\includegraphics[width=#1\linewidth]{#2}};
	\draw (#4, #5) node[fill=white,inner sep=1pt]  {\textbf{#3}};
	\end{tikzpicture}}

\title{Maximum Likelihood Uncertainty Estimation: Robustness to Outliers}
\author {
    Deebul S. Nair,\textsuperscript{\rm 1}
    Nico Hochgeschwender, \textsuperscript{\rm 1 2}
     Miguel A. Olivares-Mendez \textsuperscript{\rm 3}
}
\affiliations {
    \textsuperscript{\rm 1}Dept. of Computer Science, Bonn-Rhein-Sieg University of
    Applied Sciences, Germany.\\
    \textsuperscript{\rm 2} German Aerospace Center (DLR),
    Institute for Software Technology, Germany.\\
     \textsuperscript{\rm 3} SpaceR Research
    Group, University of Luxembourg, Luxembourg.\\
    deebul.nair@h-brs.de, nico.hochgeschwender@h-brs.de, miguel.olivaresmendez@uni.lu
}

\begin{document}

\maketitle

\begin{abstract}
 We benchmark the robustness of maximum likelihood based uncertainty estimation methods to outliers in training data for regression tasks. Outliers or noisy labels in training data results in degraded performances as well as incorrect estimation of uncertainty. We propose the use of a heavy-tailed distribution (Laplace distribution) to improve the robustness to outliers. This property is evaluated using standard regression benchmarks and on a high-dimensional regression task of monocular depth estimation, both containing outliers. In particular, heavy-tailed distribution based maximum likelihood provides better uncertainty estimates, better separation in uncertainty for out-of-distribution data, as well as better detection of adversarial attacks in the presence of outliers.
\end{abstract}

\section{Introduction}
\label{introduction}

The ability to estimate the uncertainty along with network prediction has become a relevant feature for the adoption of deep neural networks (DNN) in safety-critical and autonomous systems ~\cite{borg_safely_2019, schwalbe_survey_2020}. Accurate and calibrated uncertainties can be used to gain confidence for making decisions in autonomous systems \cite{jha_safe_2018, serban_towards_2020}. 
Uncertainty estimation is a challenging problem, especially in high dimensional data because of the lack of ground truth regarding uncertainty. 
The presence of noisy labels or outliers in the training data elevates the challenge of uncertainty estimation.
Robust uncertainty estimation is the capability of learning algorithms to correctly learn uncertainty by ignoring the outliers. In this work we focus on the following research question: \textit{For a regression problem, given a training dataset, where an $\eta$-fraction of data are outliers, how can we robustly estimate uncertainty in the predictions?}
\begin{figure}
	\centering
	\begin{tabular}{cccc}
		\begin{sideways}
			\centering
			No Outliers 
		\end{sideways}
		&\pic[0.25]	{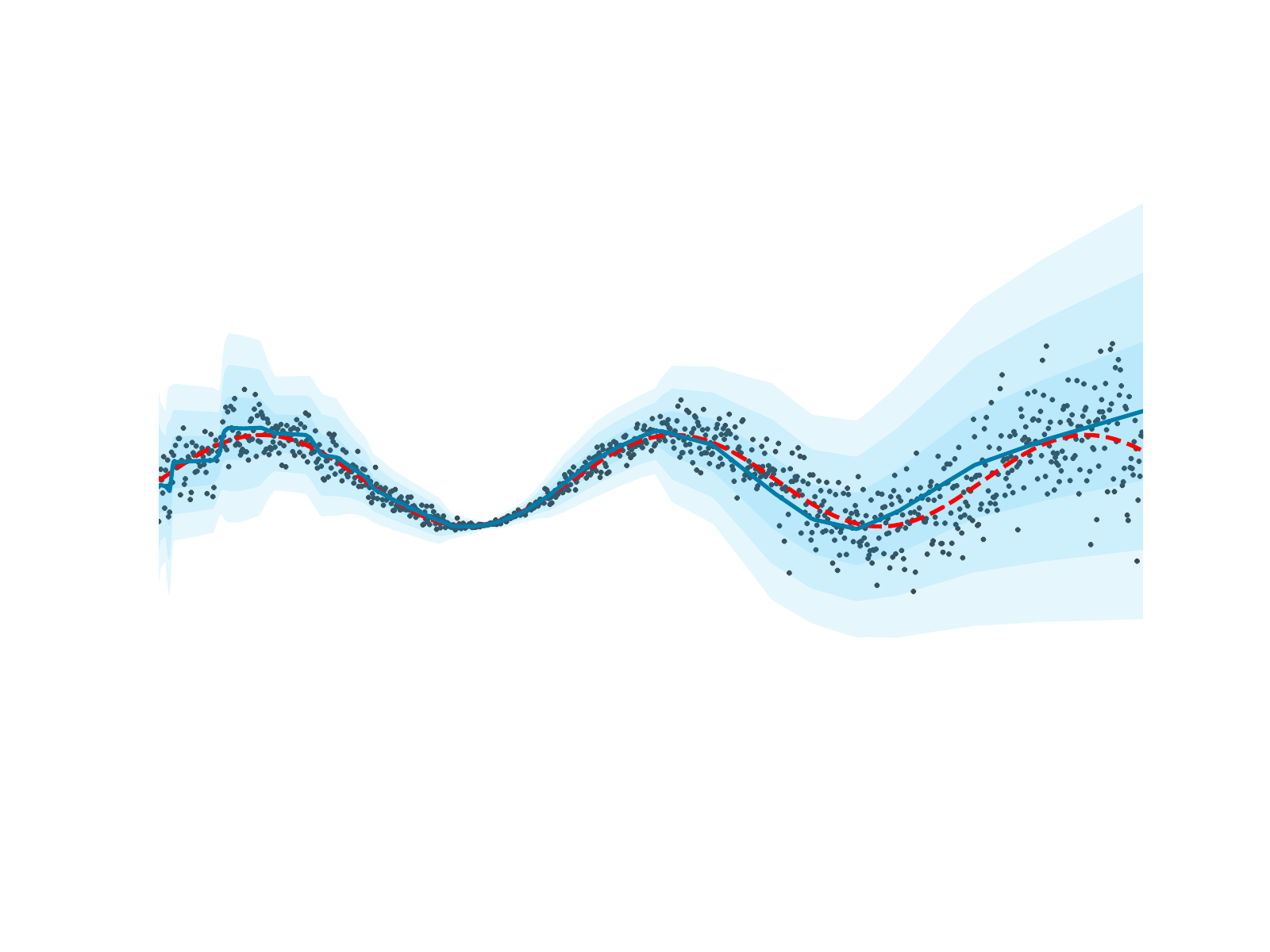}
		&\pic[0.25]{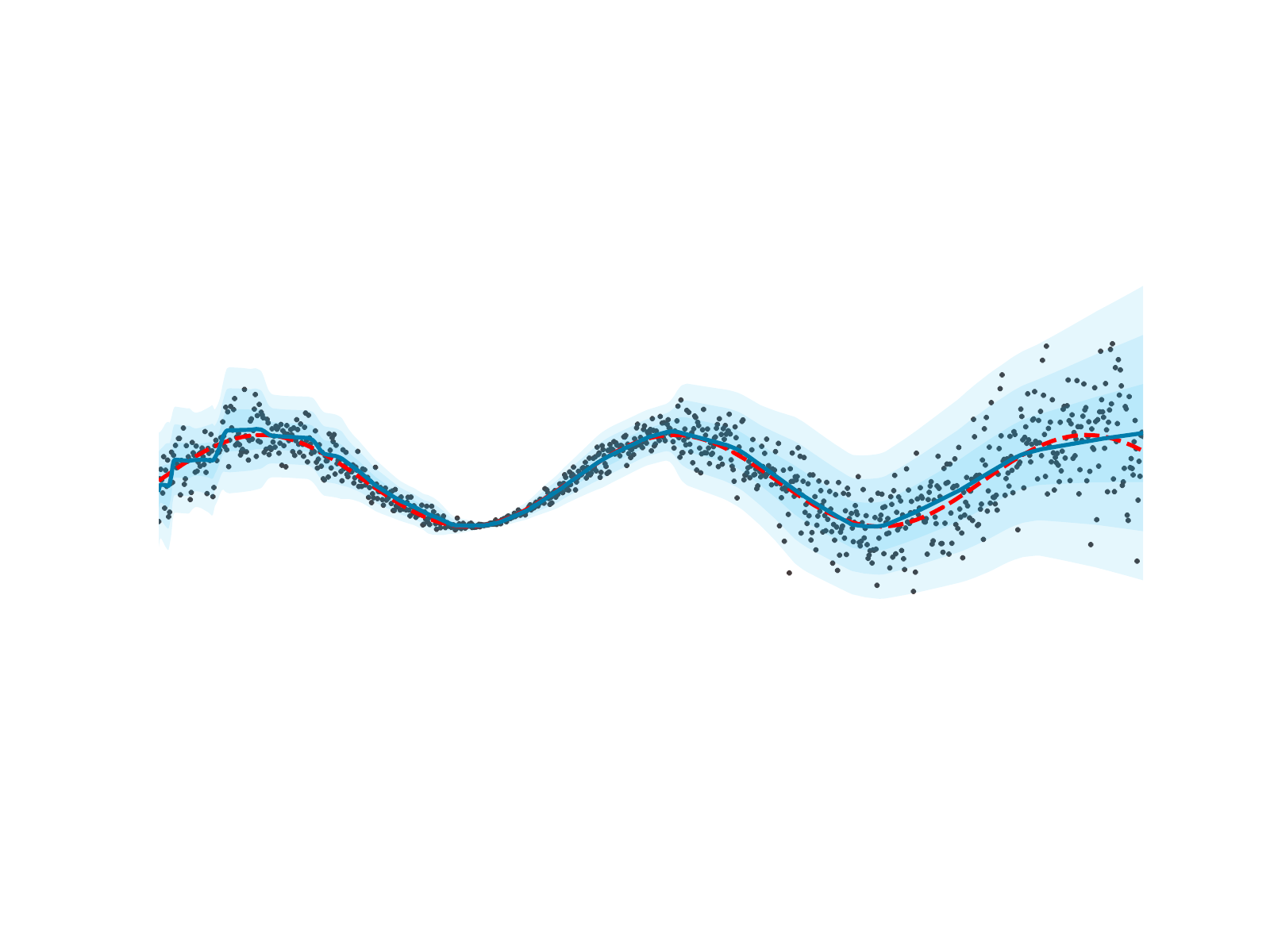}
		&\pic[0.25]	{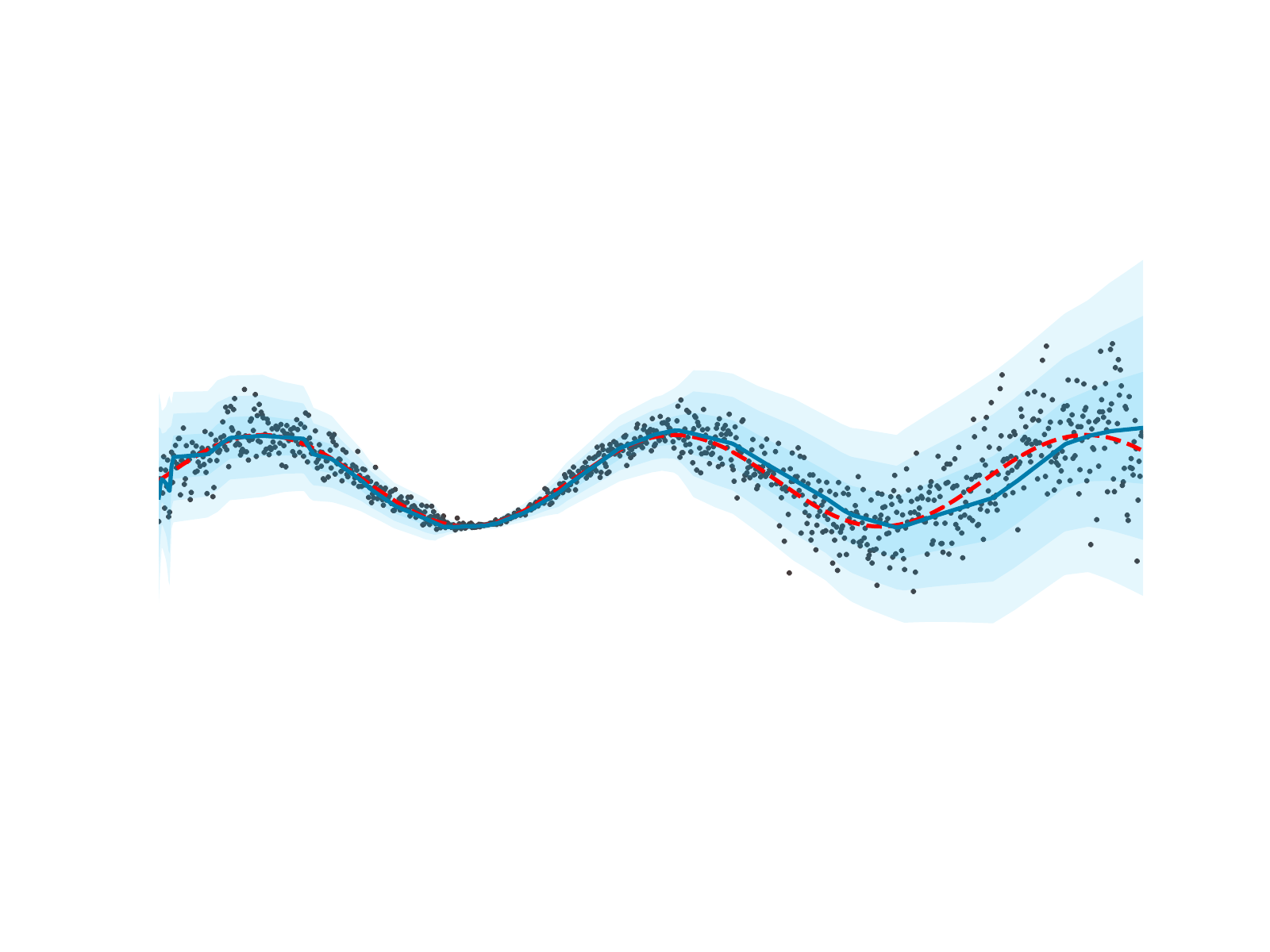}
		\\
		
		\begin{sideways}
			10\%  Outliers 
		\end{sideways}
		&\pic[0.25]{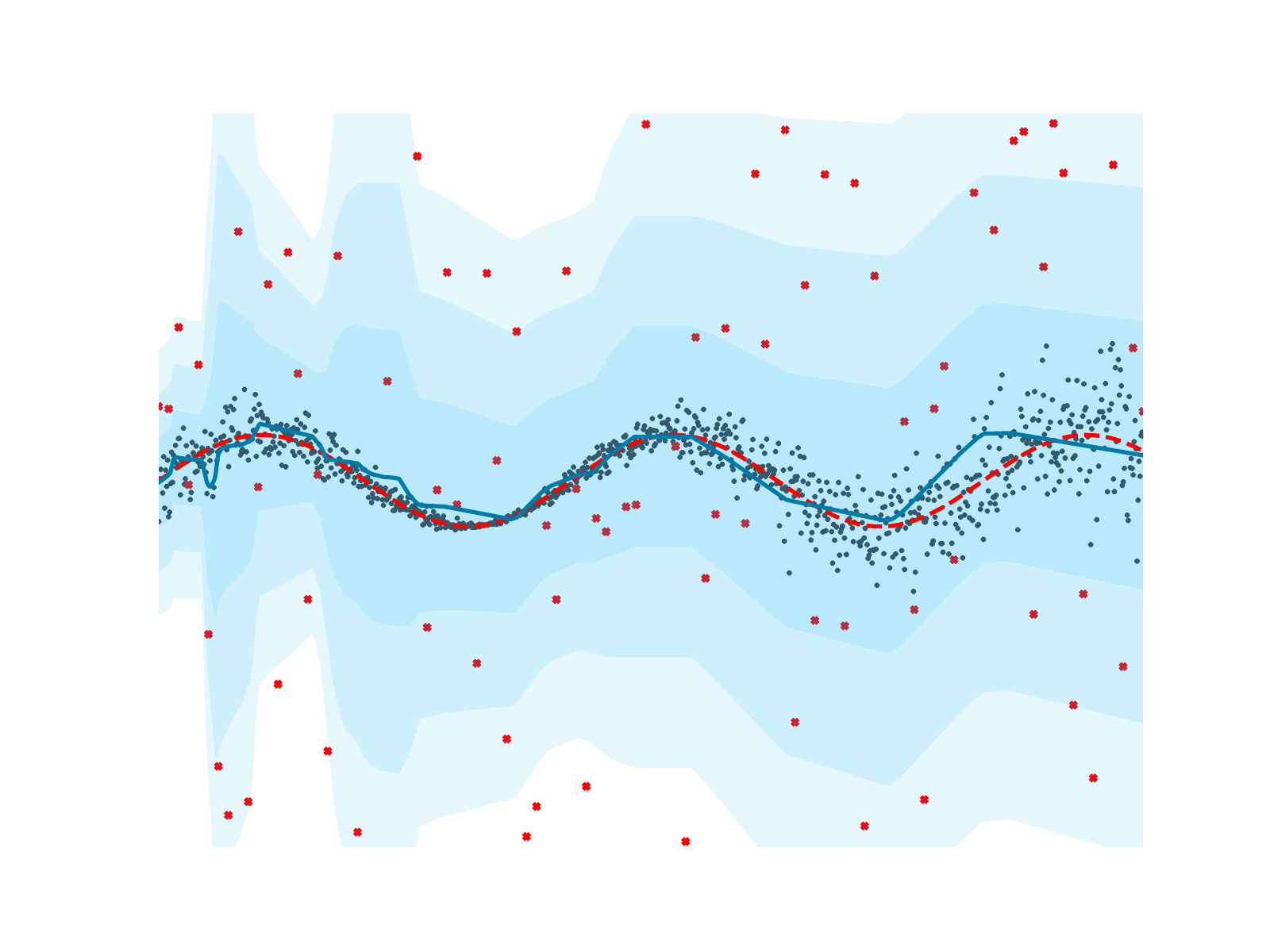}
		&\pic[0.25]{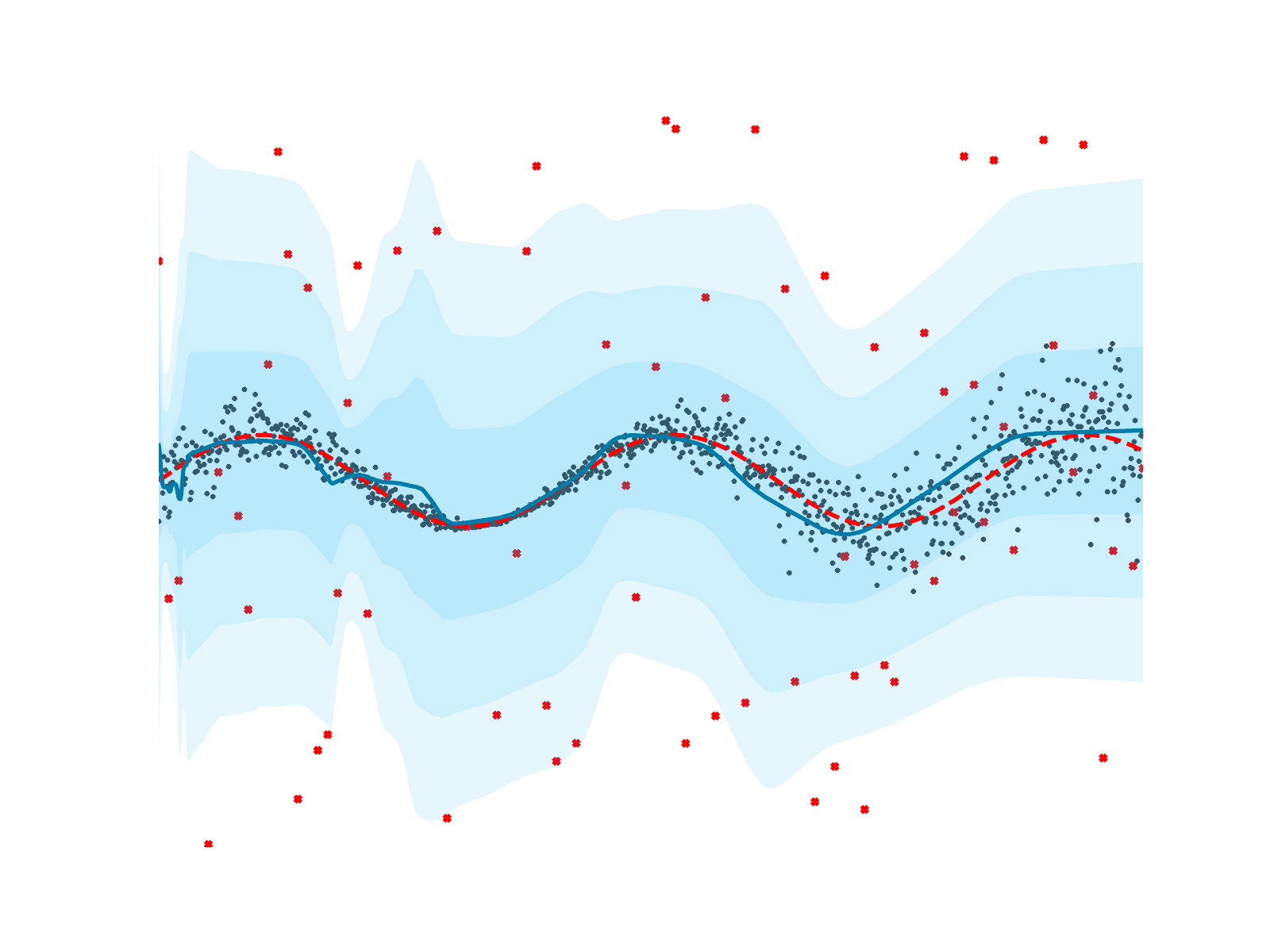}
		&\pic[0.25]{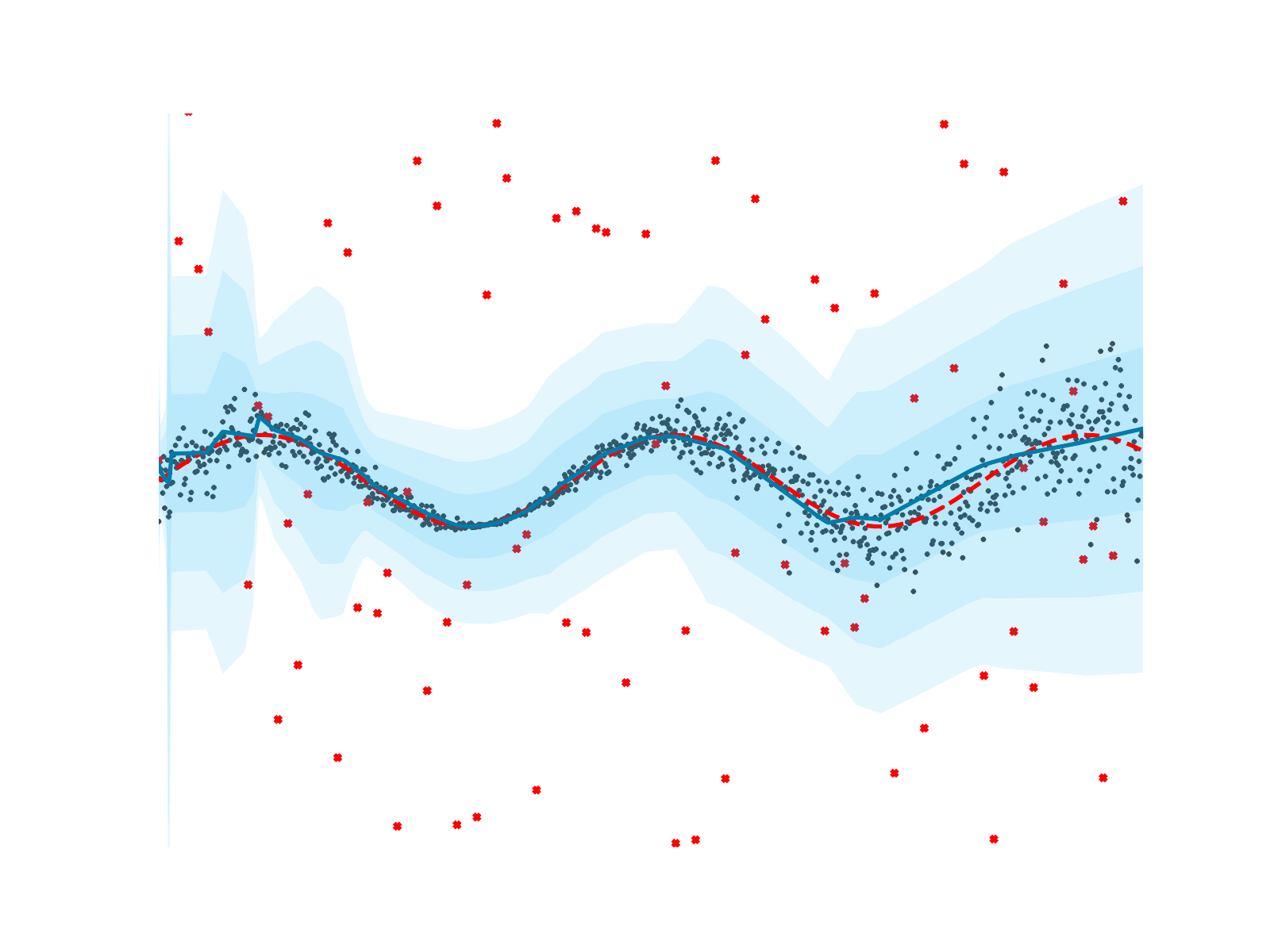}
		\\
		&(a) Gaussian &(b) Ensemble &(c) Laplace \\
		
	\end{tabular}	
	
	\caption{Depiction of the contrast in performance of uncertainty estimation on one dimensional regression data containing outliers. Row 1 shows data with no outliers while row 2 data contains 10\% outliers. The blue dots are clean data while red dots are outliers. The shaded blue region is the estimated uncertainty(three standard deviations). (a) Gaussian loss function \cite{nix1994estimating} (b) Ensemble method \cite{lakshminarayanan_simple_2016} (c) Laplace loss function (ours)} 
	\label{fig:toy-introduction}
\end{figure}

Uncertainty estimation approaches can be broadly classified in three categories: \begin{inparaenum}
	\item Bayesian methods \cite{kendall_what_2017, dusenberry2020analyzing}; 
	\item Sampling based methods \cite{gal_dropout_2016, lakshminarayanan_simple_2016};
	\item Single-model methods \cite{nix1994estimating, amini_deep_2019, sensoy_evidential_2018}
\end{inparaenum}. Although, Bayesian and sampling based methods provide state-of-the-art  uncertainty estimates \cite{dusenberry2020efficient, wen2020batchensemble} they have come at a cost: increased training time, increased prediction time and increased memory requirements. For example, the Bayesian neural learning method by converting weights to distributions \cite{blundell2015weight} requires a doubling in parameters, MC-Dropout \cite{gal_dropout_2016} with 50 Monte Carlo samples will require 50x forward passes and Deep ensembles \cite{lakshminarayanan_simple_2016} with an ensemble size of 5 will require 5x the number of weights. These become an operational challenge for their adoption in real-time autonomous systems. On the contrary, single-model based uncertainty estimation methods predict the uncertainty with a single network, no increase in weights and a single forward pass \cite{nix1994estimating, malinin_regression_2020, sensoy_evidential_2018}. In this work, we focus on single-model methods because of their applicability to autonomous systems.

All the uncertainty estimation methods have worked exclusively with clean datasets, however getting clean annotated datasets in real world applications is a difficult task. As a result, real world datasets collected even after human annotation contain large label noise \cite{AlganU21}. Learning in the presence of noisy labels is a well studied problem for the classification task \cite{arazo2019unsupervised, xia2019anchor, li2020gradient}. Also learning uncertainty in addition to the classification in the presence of label noise is now gaining importance because of safety \cite{NeverovaNV19, goel2021robustness, northcutt2021confident}.
Although all these methods have been evaluated for their prediction and uncertainty quantification for the classification task, their performance for the regression task in the presence of outliers (noisy labels) is not a well studied problem. In this work, we focus on the problem of robustly learning accurate uncertainty in the presence of outliers.

To handle outliers in statistics, heavy-tailed distribution (Student-t distribution, mixture of Gaussians and Laplace) has been introduced \cite{Lange1989RobustSM, Tak2019RobustAA, JylnkiPasi2011RobustGP}. They model the noise in the data using a heavy-tailed distribution. As the heavy-tailed distribution assumes some data can be present away from the mean, the outlier data produces comparatively less loss and have minimum impact while training.
In this paper, we study the use of heavy-tailed distributions for maximum likelihood based uncertainty estimation and benchmark their robustness to outliers in labels.  In particular, we use the Laplace distribution as a heavy-tailed distribution because of its high breakaway point \cite{bosse2016robust}, indicating the high proportion of outlier data it can handle. 

We formulate the loss function by taking the negative log-likelihood of the Laplace distribution. This loss function provides protection against outliers by reducing their effects while training. This can be demonstrated with the help of a toy example as shown in  \cref{fig:toy-introduction}. The dataset is a one dimensional (1D) regression problem where the ground truth $y$ is a sine-wave. The dataset contains heteroskedatic noise with varying levels of uncertainty across $x$. In addition to the noisy data we also add 10\% outliers to the data. The addition of outliers results in a complete mis-calibration of uncertainty estimated by Gaussian loss function \cite{nix1994estimating} and Evidential \cite{amini_deep_2019}  however, the proposed Laplace loss function uncertainty estimates are not that degraded.

We benchmark the robustness of uncertainty estimation methods when trained with noisy data. Specifically, this work makes the following contributions:
\begin{compactenum}
	\item we improve the robustness of uncertainty estimation by modeling the loss function using a heavy-tailed distribution;\par
	\item we evaluate the robustness of the loss function on a standard regression datasets benchmark and complex vision regression task of depth estimation; and\par
	\item we evaluate the use of predicted uncertainty for out-of-distribution (OOD) detection and adversarial attacks detection.\par
\end{compactenum}

\section{Uncertainty Estimation in Regression}
\label{sec:hints}

In a regression learning problem we have a training dataset $D= (x,y)$ drawn from a joint distribution $D(X,Y)$, where for each co-variate sample pair, $x \in X $ is the input data and $y \in Y $ is the label data. In addition we define a neural network $f_\theta$ parameterized by $\theta$. We train the parameters $\theta$ of the neural network by minimizing the empirical loss function $\ell$.  
\begin{equation*}
\hat{\theta} = \underset{\theta}{\text{minimize}} R(\theta)  ;\quad  R(\theta) = \underset{<x,y> \in D}{E} [\ell (x,y,\theta)]
\end{equation*}
In a typical (i.e. deterministic and non-robust) regression problem, the loss function can be the sum of squared residuals.
\begin{equation*} 
\ell(x,y, \theta) = \frac{1}{2}(y - f_\theta(x))^2
 \end{equation*}   
The above problem formulation can only do point estimates and can be considered as estimating the mean of a probability distribution.  A simple extension is to estimate the entire conditional probability distribution by predicting variance $\sigma^2$ (in case of Gaussian distribution assumption) in addition to the point estimate $\mu$ \cite{nix1994estimating}. However, a Gaussian distribution is not robust to outliers in training data.

\section{Improving Robustness with Heavy-Tailed Distribution}
\label{sec:robustness}

\begin{figure}[t]
	\centering
	\includegraphics[width=\linewidth]{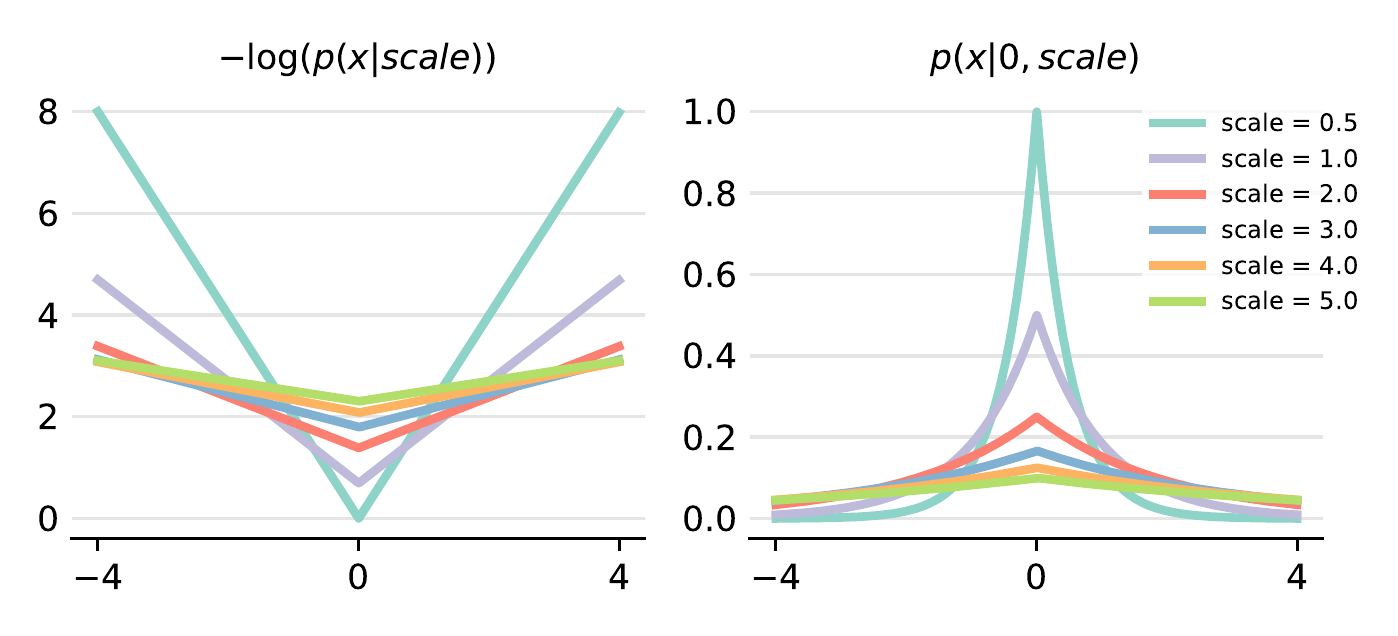}
	\caption{Negative log likelihood (NLL) (left) and probability density function (PDF) (right) of the Laplace loss function}
	\label{fig:nlllaplace}
\end{figure}

We define the problem of robust uncertainty estimation.
\begin{definition}[Robustly learning the output of a neural network with bounded second moments under additive outliers]
	Let $D$ be the set of $n$ samples so that $D = D_{clean} \cup D_{outlier}$, where $D_{clean}$ is a subset of $D$ and $D_{outlier}$ satisfies $|D_{outlier}| < \eta n $. Given an $\eta$-outlier set of samples from $D(x,y)$ and learn the output of a neural network $\hat{y}$ and $var$ such that it minimizes $|y_{clean} - \hat{y}|$  and $|var(y_{clean}) - var(\hat{y})|$
\end{definition}

Thus the goal of a robust loss function will be to accurately predict the output and its uncertainty by ignoring outliers in the data. In this paper we have selected a heavy tail distribution, Laplace distribution for modeling the uncertainty in the output.

Regression models using heavy-tailed error distributions to accommodate outliers are studied in statistics by \cite{west1984outlier}. The heavy tail of the distribution reduces the impact of the outlier while finding the maximum likelihood and thus has limited impact on the predicted value and the uncertainty. 
We choose Laplace as the heavy-tailed distribution because of its well-defined moments and it has a concave loss function, as is required for the uncertainty estimation.

\subsubsection{Laplace Maximum Likelihood}
The probability density function of a Laplace distribution is given by:
\begin{equation}
p(y| \mu, s) = \frac{1}{2s}\exp(- \frac{|y - \mu|}{s}) 
\end{equation}
where $p(y| \mu, s)$ is only defined for $s > 0$  and $\mu \in R$. We use the Negative Log likelihood (NLL) of the Laplacian distribution  $-log(p(.|\mu,s))$ for training the neural networks
\begin{equation}
\ell_{NLL}(x, y, \theta) = -\log(p(y|f_\theta^{\mu}, f_\theta^{s}) = \log(2f_\theta^{s}) + \frac{|y - f_\theta^{\mu}|}{f_\theta^{s}}
\end{equation}
The probability density function (PDF) and the NLL for different values of $f_\theta^{s}$ are plotted in \cref{fig:nlllaplace}. The loss function has several properties that makes it suitable for gradient-based optimization. The loss function is a convex function. The convexity ensures that the loss is minimum when $|y - \mu|$ is minimum and increases monotonically with respect to $|y -\mu|$ .
\begin{equation}
min(-\log(p(y|f_\theta^{\mu}, f_\theta^{s}))) \Rightarrow f_\theta^{\mu} = y \quad \quad \frac{\partial \ell}{\partial |y - f_\theta^{\mu}|} \geq 0
\end{equation}
The loss monotonically increases with respect to $f_\theta^{s}$:
\begin{equation}
\frac{\partial \ell}{\partial f_\theta^{s}} (|y - f_\theta^{\mu}|, f_\theta^{s}) \geq 0 
\end{equation}
This property helps to learn the correct aleatoric uncertainty. 
The monotonicity guaranteed by the loss function with respect to $s$ forces the optimization algorithm to reduce $f_\theta^{s}$ when there is less aleatoric noise in the data. 
As can be observed in \cref{fig:nlllaplace}, when $f_\theta^{s}$ is reduced the minimum of NLL reduces, thus ensuring that the correct uncertainties are learned.

\section{Related Work}
\label{sec:relatedwork}
\subsubsection*{Uncertainty Estimation}
\citet{nix1994estimating} estimated the uncertainty for regression problems by modeling the output of the neural network as a Gaussian distribution and learning the output using a Gaussian NLL loss function. This was the first demonstration of how the neural network can learn the uncertainty in addition to the predicted value. In the deep learning era, uncertainty estimation in large-scale vision tasks was demonstrated by \citet{blundell2015weight} which represented the weights of network with distribution. \citet{gal_dropout_2016} estimated the uncertainty by using dropouts and multiple forward passes. The current state-of-the-art in uncertainty estimation is based on ensemble methods \cite{lakshminarayanan_simple_2016, dusenberry2020efficient, wen2020batchensemble} which use a set of models under a single one. Even though all these methods provide good uncertainty estimates, they still require multiple forward passes from the model. 
\subsubsection*{Single-Model Approaches to Uncertainty Estimation}
Uncertainty estimation using a single-model can be achieved by replacing loss function based approaches \cite{malinin2018predictive, sensoy_evidential_2018, amini_deep_2019}, computing closed-form posterior for the output layer \cite{Riquelme2018Deep, SnoekRSKSSPPA15}, changing the output layer \cite{calandra2016manifold, tagasovska2019single}, spectral normalization \cite{LiuLPTBL20}, or by two-sided gradient penalty \cite{van2020uncertainty}. Laplace maximum likelihood builds on these approaches by replacing the loss function and changing the output layer.
\subsubsection*{Robust Training in Neural Networks}
Robust training in the presence of outliers is a classical statistics problem and is dominated by M-estimator methods by \citet{huber2004robust}.
In neural networks, a generalized M-estimator loss function was proposed by \citet{barron2019general}, which uses the negative log of the density function to improve robustness while training.
 \Citet{lathuiliere2018deepgum} proposed the use of Gaussian-uniform mixture model as a loss function which continuously adapts as per the outliers in data.

\section{Experiments}
In this section we benchmark performance of the proposed loss function. We will show that the proposed loss function improves robustness to outliers in data. These benchmarks are not intended to represent the state-of-the-art for any particular task; on the contrary they are intended to demonstrate the capability of our loss function when learning uncertainty with data containing outliers. We compare the following uncertainty estimation methods: 
\begin{inparaenum}
	\item \textbf{Gaussian} refers to the model introduced in \cite{nix1994estimating} 
	\item \textbf{Ensemble} corresponds to the ensemble of deep learning method proposed by \cite{lakshminarayanan_simple_2016} 
	\item \textbf{Evidential} refers to the Normal-inverse-Gamma distribution based uncertainty estimation by \cite{amini_deep_2019}
	\item \textbf{Laplace} refers to the method we propose. 
\end{inparaenum}
We benchmark these methods in terms of uncertainty prediction in a 1D outlier regression dataset, real world regression datasets and NyuV2 monocular depth estimation dataset with outliers \cite{hutchison_indoor_2012}. 

We first discuss the need for new metrics for uncertainty comparison and select an appropriate metric to benchmark the predicted uncertainty. In the subsequent section we focus on benchmarking the breakaway point, which is the percentage of outliers that can be present in the data without any significant change in the predicted output. We use a synthetic 1D dataset for adding different percentage of outliers. As we will show, our proposed loss function is particularly effective even with a high percentage of outliers. We further benchmark the methods on real world datasets of regression to demonstrate how the proposed loss function performs compared to other methods. Finally, we focus on vision based learning task of monocular depth estimation. Here we test the robustness of the methods to outliers by using the original depth dataset with outlier sensor noise. The results demonstrate the capability of our proposed loss function to learn correct depth even in the presence of outliers.

\begin{figure*}[t]
	\igwc{0.51}{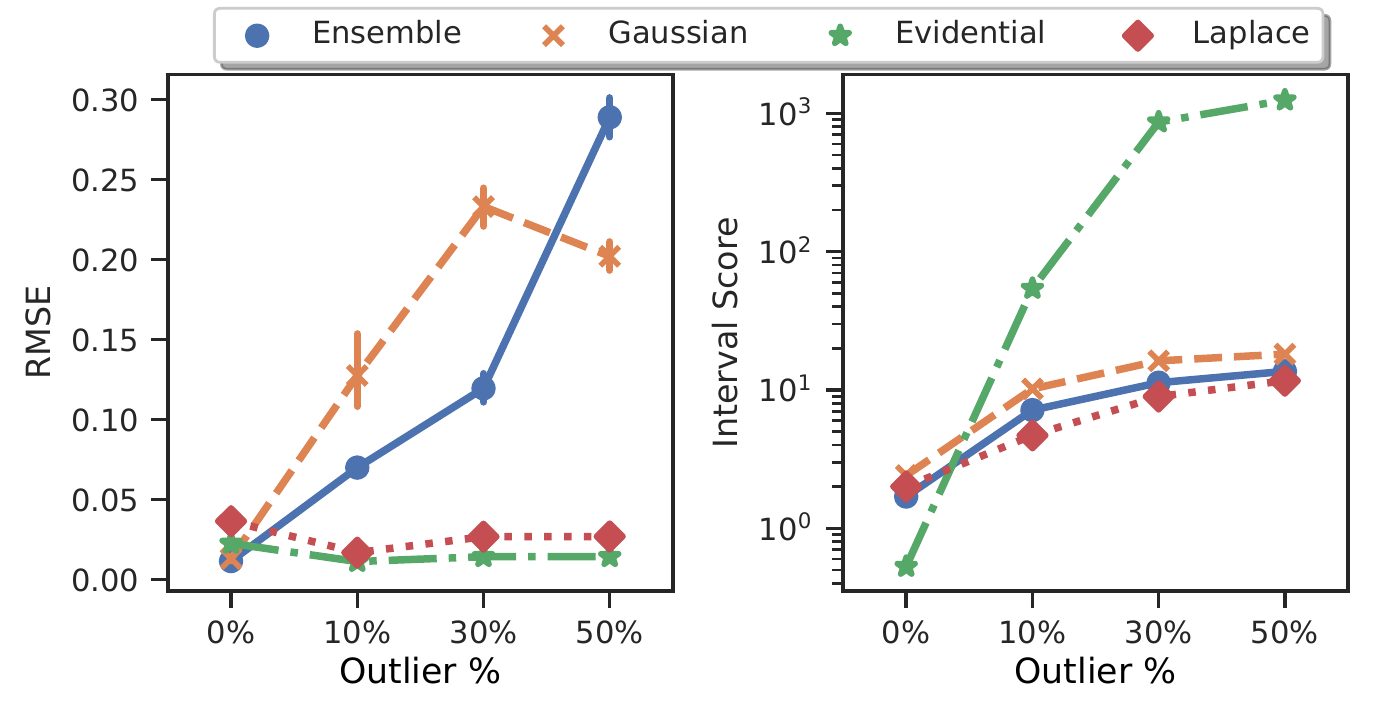}{\textbf{A}}{-3.5}{2.6}
	\igwc{0.49}{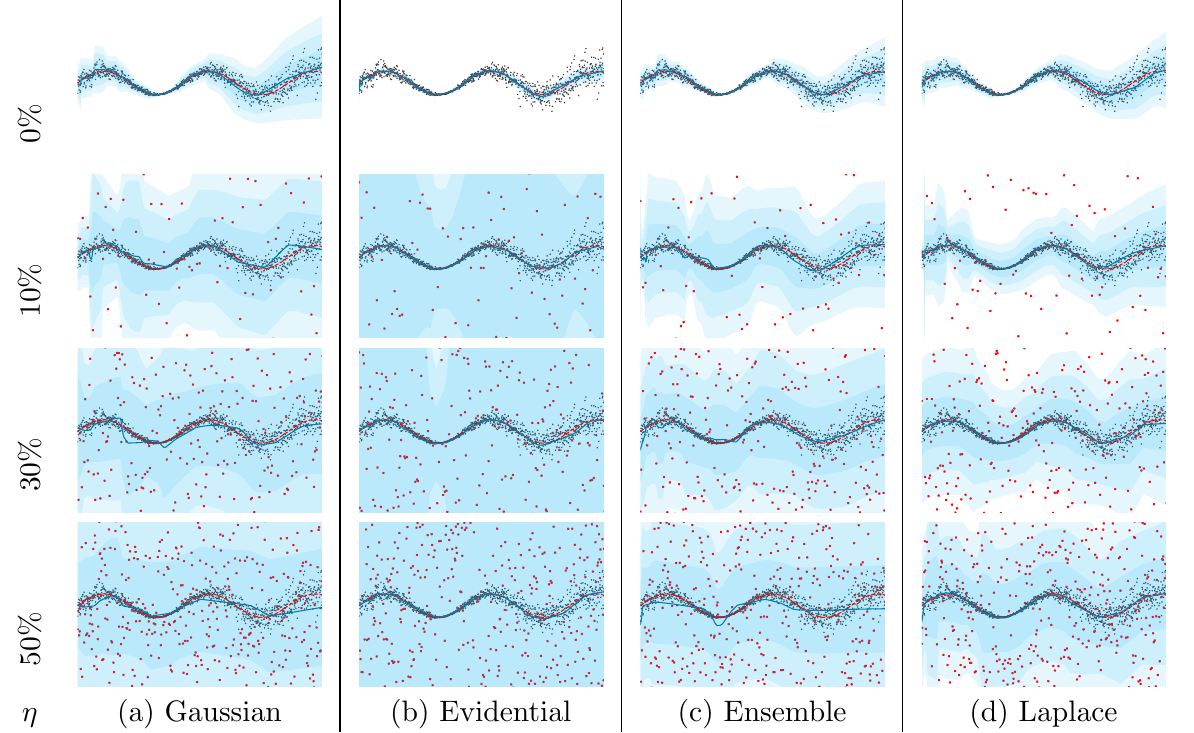}{\textbf{B}}{-4.1}{2.4}
	\caption{\textbf{Breakaway Point estimation}: \textbf{(A)}  The change in RMSE and Interval score with respect to outlier percentage in the data for the 1D regression dataset. \textbf{(B)}: The uncertainty of the neural network output with different methods learned at different outlier percentages $\eta$. The shaded blue region is three standard deviations. 
		\label{fig:breakawayExperiment}
	}
\end{figure*}
\subsection{Assessing Uncertainty Estimation Techniques}
As this work focuses on the robustness capacity of DNNs to correctly estimate the uncertainty, we needed metrics which access the quality of the estimated uncertainty separately from the quality of prediction. \emph{Scoring Rules} are a group of such metrics which assess the quality of predicted uncertainties, by assigning a numerical score based on both the prediction and uncertainty \cite{gneiting2007strictly}. 
Most of the literature in uncertainty estimation for regression use root mean square error (RMSE) for comparing performance of model and NLL for comparing performance of uncertainty. 
Although NLL is a proper Scoring Rule it is not comparable across different distributions. 
In this work we use \emph{Interval score} Scoring Rule. 
\subsubsection{Interval score} uses the prediction interval, with lower and upper endpoints represented by predictive quantiles at levels $\frac{\alpha}{2}$ and $1 - \frac{\alpha}{2}$. The Interval score is defined as :
\begin{equation}
S_\alpha^{int} (l,u;y) = (u - l) +\frac{2}{\alpha}(l-y)\mathbf{1}\{y<l\} + \frac{2}{\alpha}(y-u)\mathbf{1}\{y>u\}
\end{equation}
The score rewards narrow predictions and penalizes prediction outside the interval. For our experiments we fix the $\alpha$ to 95\%. Based on the mean and variance predicted by the neural network, we first calculate the 95\% quantile prediction interval ($l$, $u$) which is then used to calculate the Interval score. 
As the Interval score compares the 95\% quantile prediction interval it becomes comparable across different distributions (in our case Gaussian and Laplace distribution).

\subsection{Empirical Breakaway Point Benchmark }
In this experiment, we empirically benchmark the breakaway point for all the methods. Breakaway point is the percentage of outliers that can be present in the dataset after which the estimator predicts statistically wrong outputs \cite{huber2004robust}. Here we interpret the outputs as both the predicted output and the predicted uncertainty. The regression problem we consider is the 1D regression problem as shown in \cref{fig:toy-introduction} which is a sine wave with increasing aleatoric uncertainties along the x-axis. The particular dataset was selected such that the network has to learn the underlying aleatoric uncertainty even in the presence of outliers. The neural network is 4 layer fully connected network with 100 neurons in each layer with rectified linear unit (ReLU) activation function. An illustrative scatter plot of the dataset with outliers and the predicted uncertainty is shown in \cref{fig:breakawayExperiment}-B. In this benchmark we start with no outliers and increase the outliers upto 50\% of the original dataset and record RMSE and Interval score for each dataset. \cref{fig:breakawayExperiment}-A plots the RMSE and Interval score for different levels of outliers. Gaussian and Ensemble methods have breakaway points at 10\% outliers for the output prediction (\cref{fig:breakawayExperiment}-A left plot), while for the uncertainties the breakaway point is 10\% for Evidential learning (\cref{fig:breakawayExperiment}-A right plot). In this benchmark, we can conclude that the breakaway points are different for the predicted output value and its uncertainty. Some methods are better at output prediction and worse at uncertainty prediction, while Laplace method performs best on both the scenarios in the given example. 


\begin{figure}
	\centering
	\includegraphics[width=\linewidth,trim={0.5cm 0 1.3cm 0},clip]{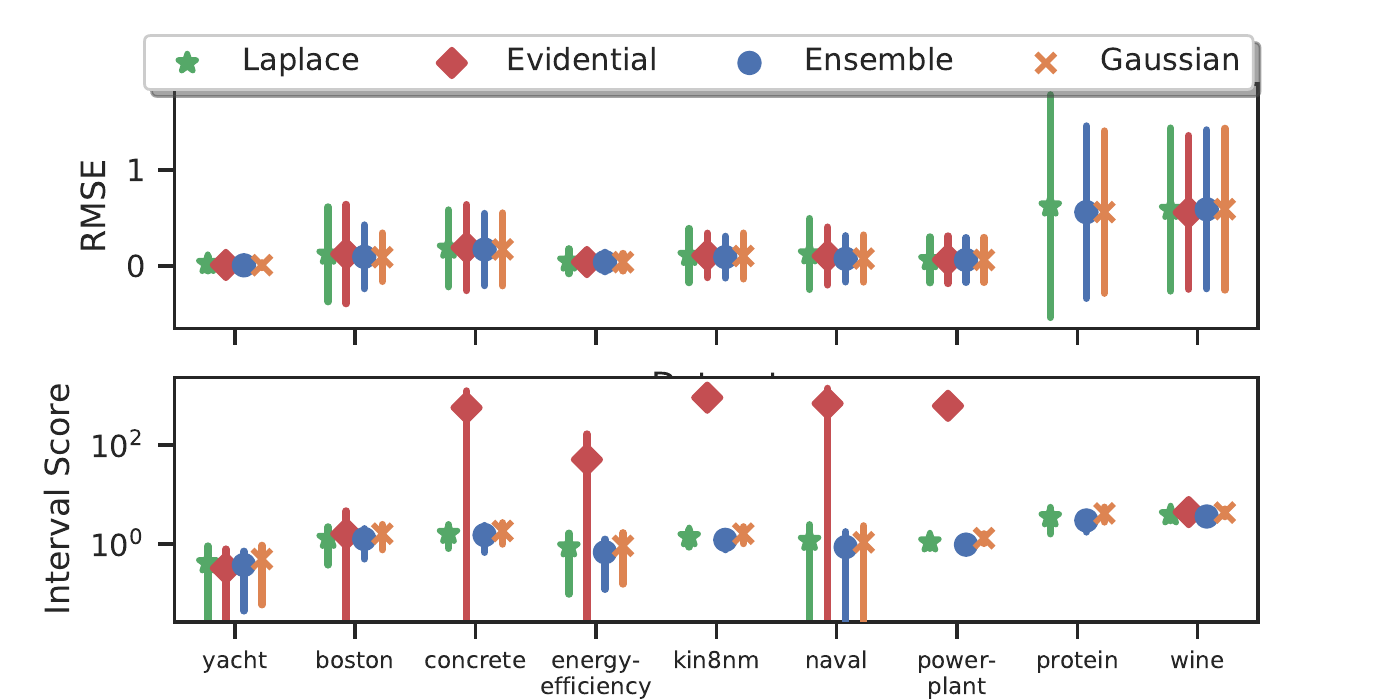}
	\caption{\textbf{Benchmark Regression Tasks}: RMSE and Interval score for different methods and datasets. Our method achieves statistical comparable results in eight of the nine datasets in RMSE. While for Interval score it provides comparable in all the nine datasets}
	\label{fig:uci}
\end{figure}

\begin{figure*}	
	\centering
	\igwc{0.29}{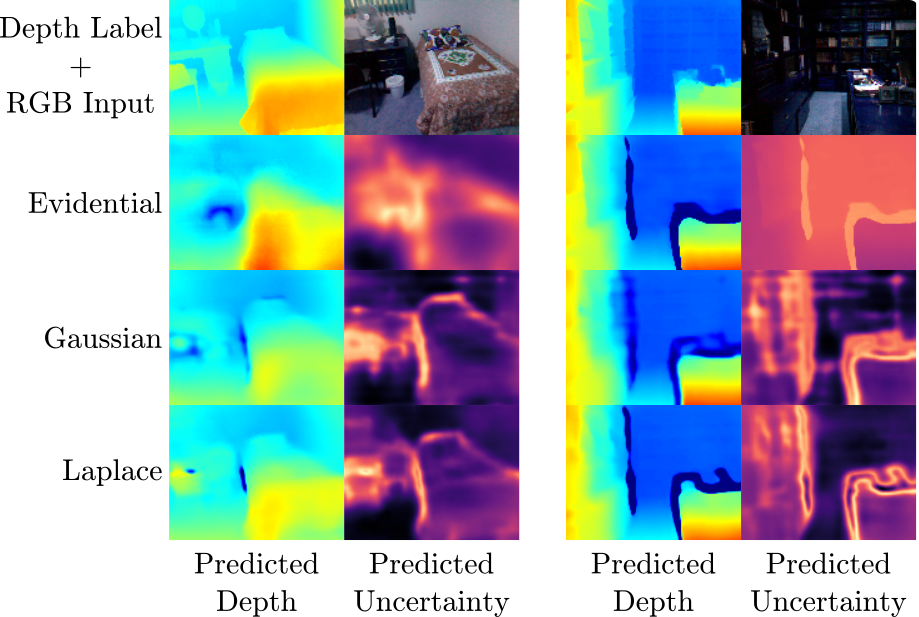}{\textbf{A}}{-2.5}{1.9}
	\igwc{0.22}{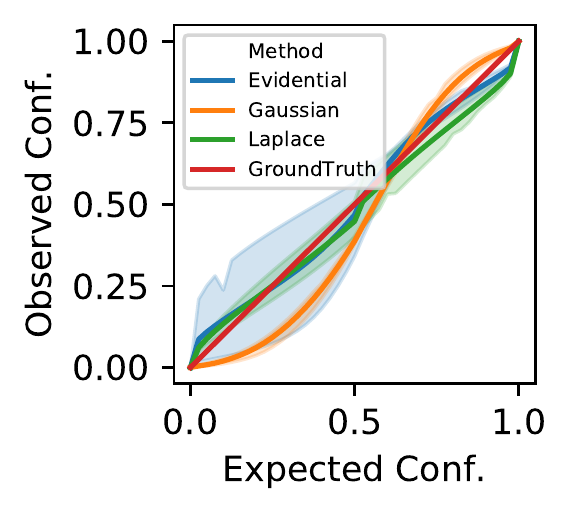}{\textbf{B}}{-1.5}{1.9}
	\igwc{0.47}{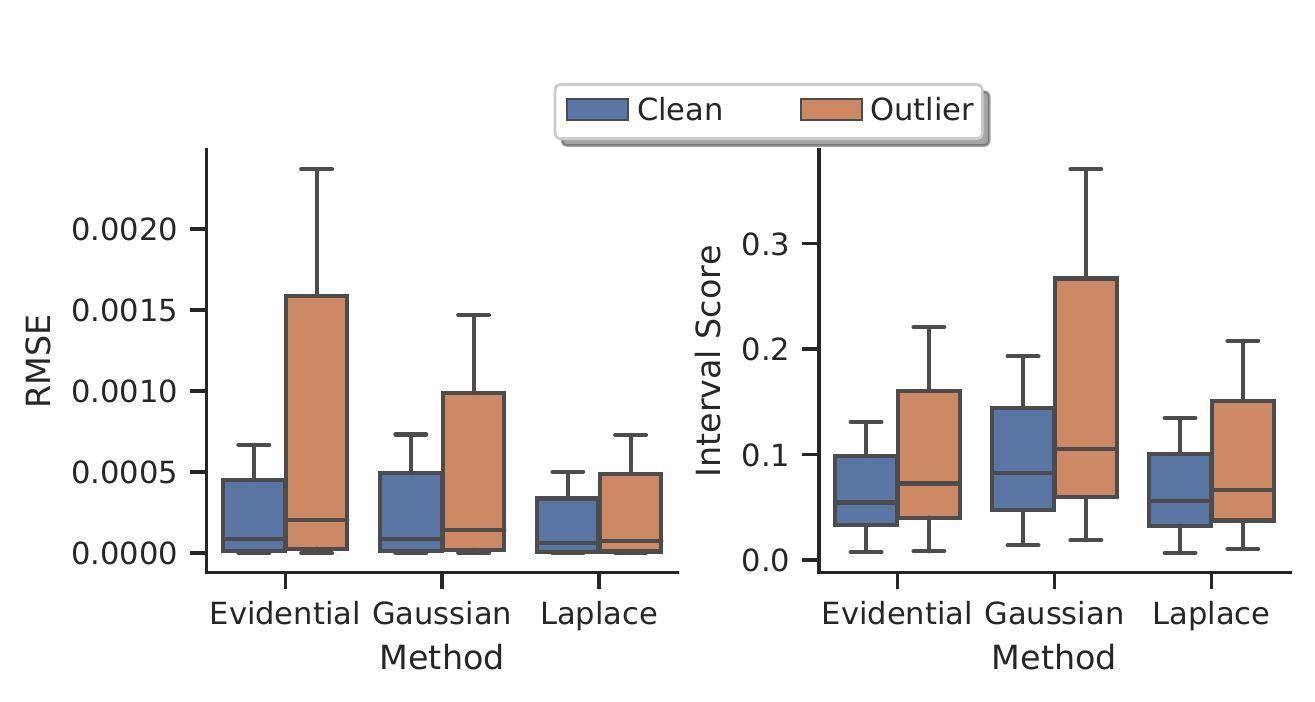}{\textbf{C}}{-3.5}{1.35}
	\caption{\textbf{Monocular Depth Estimation} Trained on the NYU-Depth-v2-dataset containing outliers. \textbf{(A)} Example of estimated depth images and corresponding uncertainty per method.\textbf{ (B)} Uncertainty calibration plot \textbf{(C)} Performance comparison between models trained on clean data and outlier data.}
	\label{fig:sampledepthimagescollage}
\end{figure*}

\subsection{Benchmark Regression Tasks}
In addition to the toy datasets we validate the uncertainty estimation with respect to real world datasets as used in the \citet{lakshminarayanan_simple_2016, gal_dropout_2016, amini_deep_2019}. We replicate the experiment and compare RMSE and the Interval score. The purpose of the experiment is to compare the uncertainty quality obtained from different methods to that of the proposed method on real world datasets. Each experiment is executed  20 times each with random sampling of the train and test dataset. The network used is a fully-connected network with 3 hidden layers containing 100 neurons, and was trained to convergence. All models were trained with learning rate $\eta = 5e^{-3}$ and batch size of 512. The results are shown in \cref{fig:uci}. As seen in the results the proposed Laplace method performs comparable to other uncertainty estimation methods both in RMSE and Interval score. The Evidential method shows some deviation in the predicted uncertainty from the other methods in the Interval score plot for all four datasets. 

\subsection{Monocular Depth Estimation with Outliers}

\begin{figure}
	\centering
	\includegraphics[width=1.0\linewidth]{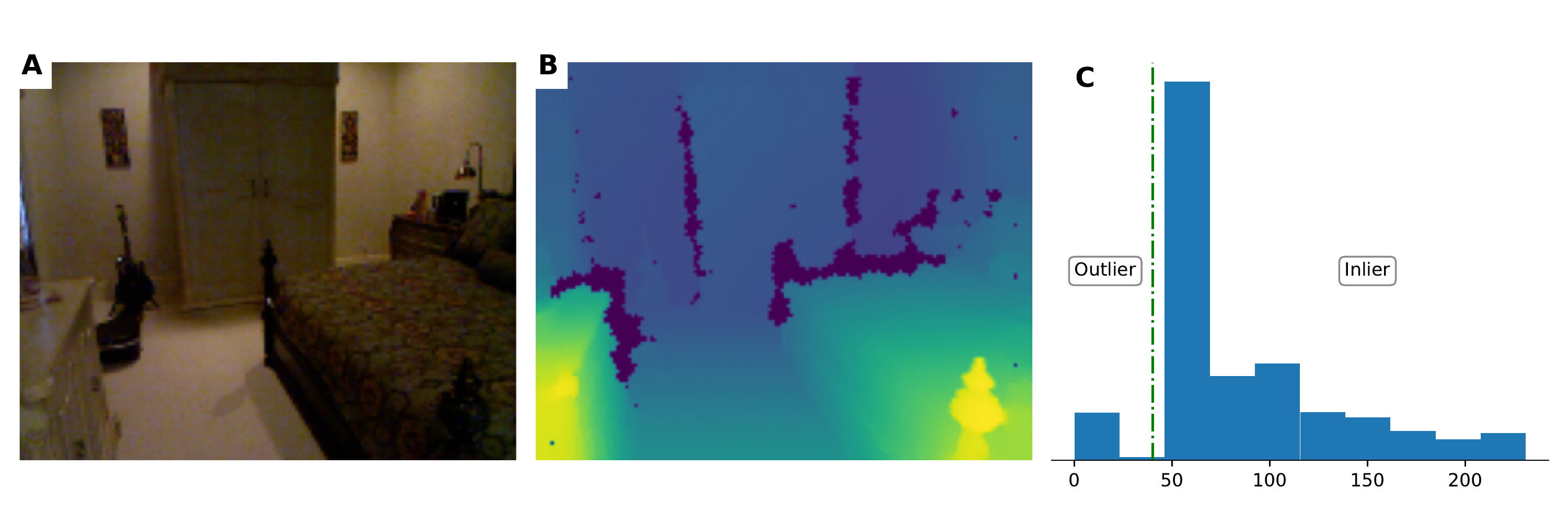}
	\caption[Outliers in Depth estimation dataset]{\textbf{Outliers in the NYU-Depth-v2-dataset \cite{hutchison_indoor_2012}}. (A) Input RGB image (B) Depth image captured from Kinect camera. The dark spots are outliers where no depth information is available because of sensor noise (C) The histogram of the depth where the outliers can be separated}
	\label{fig:outlierdepth}
\end{figure}
Monocular depth estimation is a supervised regression task, where the model is trained to predict depth of each corresponding pixel.
The input data is a RGB image of size HxWx3 and the label is a depth image with size HxW whose pixel values represent the depth of the scene in meters. 
We evaluate monocular depth estimation on the NYU-Depth-v2-dataset which is collected in an indoor environment using a Kinect v1 camera.  

In fields of robotics, the datasets collected are noisy and always contaminated with outliers. For example, in \cref{fig:outlierdepth}-B we can observe black spots (outliers) which contain no depth information, in the collected depth labels because of sensor noise in the Kinect v1 camera.
In the NYU-Depth-v2-dataset the outliers in the images were removed by using the Levin Colorization method. For our work, we have not removed the outliers and used the depth labels along-with the outliers.
In order for the results to be comparable to \citet{amini_deep_2019}, rather than directly using the outlier labels, we simulate the outliers in the cleaned depth data using simulated noise model of a Kinect v1 camera \cite{handa_etal_2014, Barron_eta_2013A, Bohg_etal_2014}. 

The training dataset consists of 27K RGB-depth image pairs. For comparison, we use the architecture similar to \citet{amini_deep_2019} which is a U-Net \cite{ronneberger2015u} with spatial dropout.  The goal of the experiment is not to prove state-of-the-art depth estimation but comparison of uncertainty estimation, hence we have limited to the popular depth estimation architecture.
The U-Net model consists of five convolutional and pooling blocks in both the down-sampling and up-sampling parts. The input image shape is (160,128).
The final layer outputs HxWx$n$ layers, where $n = 2$ for Gaussian and Laplace regression  while $n = 4$ for Evidential regression.
We perform three independent runs for all the methods and report the distribution in all the experiments. The hyper-parameters used are: Adam optimizer with learning rate $5e^{-5}$, batch size of 32 and over 70000 iterations.
The evaluation is done using a disjoint dataset with non-outlier depth labels.

\subsubsection{Clean Data vs Outlier Data}

We report the comparison of RMSE and Interval score between models trained on clean depth data and depth data containing outliers in \cref{fig:sampledepthimagescollage}-C. Clean depth data here we refer to the NYU-Depth-v2-dataset in which the outliers are removed by using the Levin Colorization method, while outlier data is the depth data containing outliers. The goal of the experiment is to showcase the degradation in the uncertainty estimation when the depth data contains outliers. In \cref{fig:sampledepthimagescollage}-C we can observe the change in performance and uncertainty estimation when trained with outliers in data. First we observe that adding outliers degrades performance of all the three methods which is as expected.
For Evidential, we observe there is large shift in RMSE but less in Interval score, while for Gaussian the Interval score shift is highest. 
The change between both the metrics is minimal for Laplace regression indicating robustness to outliers.

\subsubsection{Uncertainty Calibration} Here we measure uncertainty calibration of each method using calibration curves, where the expected confidence (inverse of uncertainty) is plotted with respect to observed confidence. The expected confidence is calculated based on the error; for samples with minimum error we expect maximum confidence and vice versa.  For a well calibrated method, the curve will follow the line $y=x$. In \cref{fig:sampledepthimagescollage}-B we plot the calibration curves for three methods. The results show calibration curves over three independent trials represented by the shaded region in different color. Here we observe that the Gaussian method underestimates confidence in the low confidence region and overestimates confidence in the high confidence region. Evidential shows calibrated predictions, however, there is large spread between indicating lack of reliability in learning. Overall Laplace has the comparative best calibration over multiple trials.

%
	

\subsubsection{Adversarial Attack Detection}
\begin{figure}
	\centering
	\igwc{0.49}{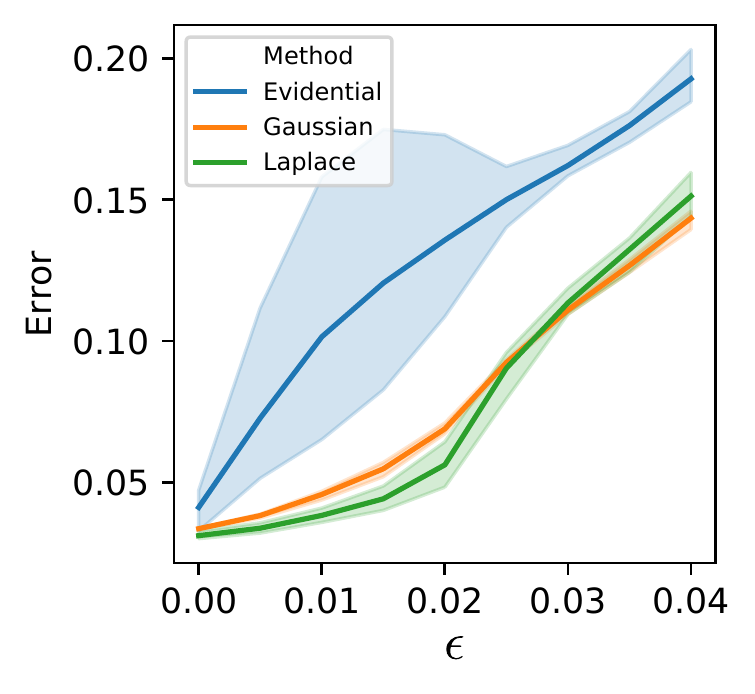}{\textbf{A}}{-1}{2}
	\igwc{0.49}{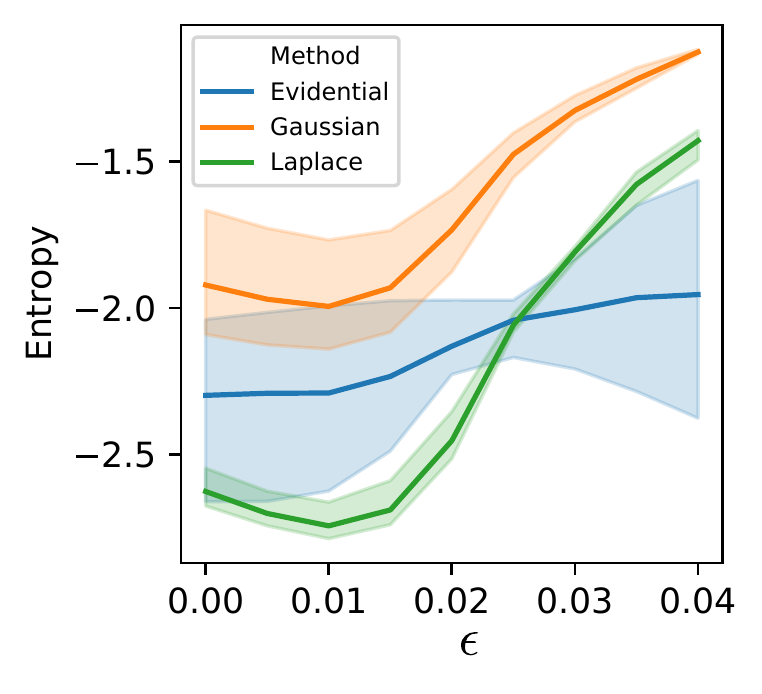}{\textbf{B}}{-1}{2}
	\caption{\textbf{Adversarial detection} (A) Error, (B) Entropy as a function to adversarial perturbation $\epsilon$ on the NYU-Depth-v2-dataset.}
	\label{fig:advmethoderror}
\end{figure}
The models which are trained in the above section are evaluated on an adversarial dataset generated using the Fast Gradient Sign Method (FGSM) \cite{goodfellow_2015}, using various values of the adversarial perturbation coefficient $\epsilon$. 
Adversarial images are artificial generated images with $\epsilon$ intentional feature perturbation which causes the model to give wrong predictions.
The adversarial attack results for different values of $\epsilon$ are shown in \cref{fig:advmethoderror}.
We observe in  \cref{fig:advmethoderror}-A, that the error increases with increase in adversarial attack for all the methods.
In \cref{fig:advmethoderror}-B, we plot the entropy over the range of $\epsilon$.
The desirable result is that the entropy should increase for increasing level of $\epsilon$  \cite{amini_deep_2019}, indicating higher uncertainty for higher adversarial inputs.
We observe Laplace and Gaussian show an increase in entropy as adversarial input increases. For Gaussian, we also observe that for lower $\epsilon$ it has higher spread of entropy, thus indicating high uncertainty for no adversarial input images. The rise of uncertainty is minimum for Evidential.
Thus we can conclude that the Laplace loss function learns to estimate appropriate uncertainty  while learning from data containing outliers.

\subsubsection{Out-of-Distribution Detection}

\begin{figure}
	\centering
	\igwc{0.5}{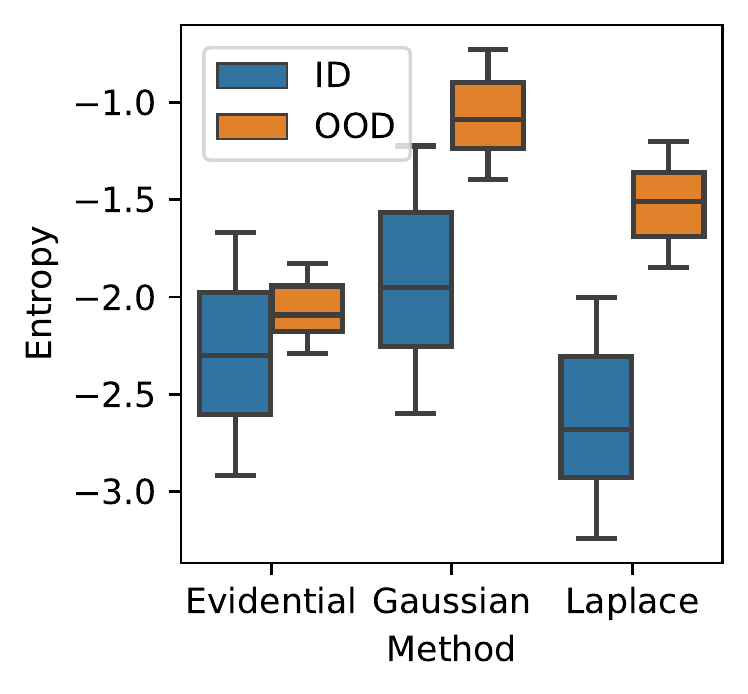}{\textbf{A}}{-1.2}{2}
	\begin{tikzpicture}
	\draw (0, 0) node[inner sep=0] {\includegraphics[trim=320 0 0 0,clip,width=0.4\linewidth]{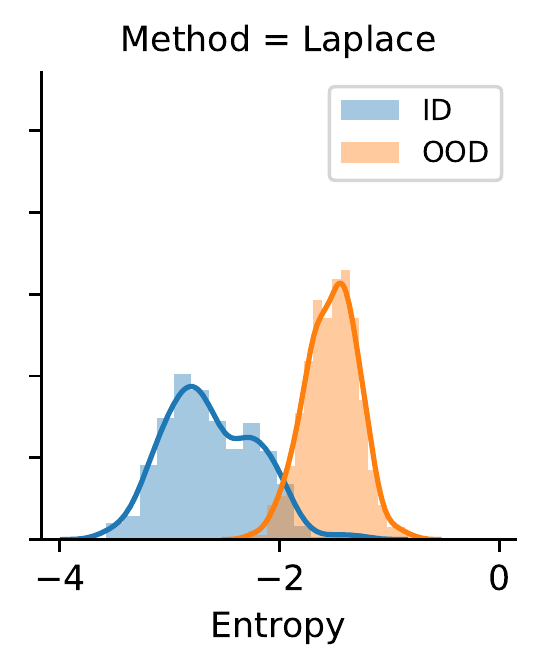}};
	\draw (-1.4, 1.9) node {\textbf{B}};
	\end{tikzpicture}
	\caption{\textbf{Out of distribution} Entropy comparison for in-distribution (ID) data and out-of-distribution (OOD) data \textbf{(A)} Comparison Box plot of entropy between ID and OOD data \textbf{(B)} Entropy distribution for the Laplace method }
	\label{fig:ood}
\end{figure}

We also evaluated our trained models on an OOD dataset. The ApolloScape dataset by \citealt{huang_2018} was selected as the OOD dataset. 
The ApolloScape dataset is also a depth estimation dataset consisting of images from outdoor settings, thus making it appropriate for testing OOD.
The desirable result is an increased uncertainty for OOD data predictions as compared to In-distribution (ID) predictions.
The predicted uncertainty for OOD data should also be statistically different than the ID predictions.
This clear separation helps in defining a threshold value for making informed decisions on trusting the predictions. 
\cref{fig:ood}-A plots the interquartile box plots of entropy which shows that only the Laplace method is able to statistically distinguish between ID and OOD data. The separation is minimum for the Evidential method.
This is corroborated in \cref{fig:ood}-B where we plot the distribution of entropy for both ID and OOD data from the Laplace method. 

\section{Conclusion and Outlook}
\label{sec:outlook}
We have benchmarked maximum likelihood based uncertainty estimation for deep regression.
We observed that using Gaussian-based loss functions are not robust to outliers in the training data and consequently provide inaccurate uncertainty estimates.
We proposed a heavy-tailed distribution based loss function as an alternative in order to improve robustness to outliers.
The heavy-tailed Laplace loss function accurately estimates uncertainty in predictions and has a high breakaway point compared to the other methods. 
 When applied to high dimensional datasets containing outliers, such as depth estimation datasets, the Laplace loss function is able to better estimate the uncertainties.
Our proposed loss function improves state-of-the-art in three uncertainty estimating benchmarks: \begin{inparaenum}
	\item high breakaway point;
	\item detection of OOD ;
	\item detection of adversarial inputs.
\end{inparaenum}
In the future work, we would benchmark performance by modeling the loss function with other heavy-tailed distributions.
The proposed robust loss function could benefit in building software which uses uncertainty from the neural network  for safe deployment of deep neural networks in autonomous systems.

\section{Acknowledgments}
Deebul Nair gratefully acknowledges the ongoing support
of the Bonn-Aachen International Center for Information Technology and a PhD
scholarship from the Graduate Institute of the Bonn-Rhein-Sieg University. This
work was supported by the European Union’s Horizon 2020 project SciRoc (grant
agreement No 780086), SESAME (grant agreement No 101017258) and DLR
CASSy project.
All the code is available at	\url{https://github.com/deebuls/uncertainty-robustness-deep-regression-benchmark},

\bibliography{nair}

\begin{thebibliography}{45}
\providecommand{\natexlab}[1]{#1}

\bibitem[{Algan and Ulusoy(2021)}]{AlganU21}
Algan, G.; and Ulusoy, I. 2021.
\newblock Image classification with deep learning in the presence of noisy
  labels: {A} survey.
\newblock \emph{Knowledge Based Systems}, 215: 106771.

\bibitem[{Amini et~al.(2020)Amini, Schwarting, Soleimany, and
  Rus}]{amini_deep_2019}
Amini, A.; Schwarting, W.; Soleimany, A.; and Rus, D. 2020.
\newblock Deep {Evidential} {Regression}.
\newblock In \emph{Advances in {Neural} {Information} {Processing} {Systems}},
  volume~33, 14927--14937.

\bibitem[{Arazo et~al.(2019)Arazo, Ortego, Albert, O’Connor, and
  McGuinness}]{arazo2019unsupervised}
Arazo, E.; Ortego, D.; Albert, P.; O’Connor, N.; and McGuinness, K. 2019.
\newblock Unsupervised label noise modeling and loss correction.
\newblock In \emph{International Conference on Machine Learning}, 312--321.

\bibitem[{Barron(2019)}]{barron2019general}
Barron, J.~T. 2019.
\newblock A general and adaptive robust loss function.
\newblock In \emph{IEEE/CVF Conference on Computer Vision and Pattern
  Recognition}, 4331--4339.

\bibitem[{Barron and Malik(2013)}]{Barron_eta_2013A}
Barron, J.~T.; and Malik, J. 2013.
\newblock Intrinsic Scene Properties from a Single RGB-D Image.
\newblock \emph{IEEE/CVF Conference on Computer Vision and Pattern
  Recognition}.

\bibitem[{Blundell et~al.(2015)Blundell, Cornebise, Kavukcuoglu, and
  Wierstra}]{blundell2015weight}
Blundell, C.; Cornebise, J.; Kavukcuoglu, K.; and Wierstra, D. 2015.
\newblock Weight uncertainty in neural network.
\newblock In \emph{International Conference on Machine Learning}, 1613--1622.

\bibitem[{Bohg et~al.(2014)Bohg, Romero, Herzog, and Schaal}]{Bohg_etal_2014}
Bohg, J.; Romero, J.; Herzog, A.; and Schaal, S. 2014.
\newblock Robot arm pose estimation through pixel-wise part classification.
\newblock \emph{International Conference on Robotics and Automation}.

\bibitem[{Borg et~al.(2019)Borg, Englund, Wnuk, Duran, Levandowski, Gao, Tan,
  Kaijser, Lönn, and Törnqvist}]{borg_safely_2019}
Borg, M.; Englund, C.; Wnuk, K.; Duran, B.; Levandowski, C.; Gao, S.; Tan, Y.;
  Kaijser, H.; Lönn, H.; and Törnqvist, J. 2019.
\newblock Safely {Entering} the {Deep}: {A} {Review} of {Verification} and
  {Validation} for {Machine} {Learning} and a {Challenge} {Elicitation} in the
  {Automotive} {Industry}.
\newblock \emph{Journal of Automotive Software Engineering}, 1(1): 1--19.

\bibitem[{Bosse, Agamennoni, and Gilitschenski(2016)}]{bosse2016robust}
Bosse, M.; Agamennoni, G.; and Gilitschenski, I. 2016.
\newblock Robust Estimation and Applications in Robotics.
\newblock \emph{Found. Trends Robotics}, 4(4): 225--269.

\bibitem[{Calandra et~al.(2016)Calandra, Peters, Rasmussen, and
  Deisenroth}]{calandra2016manifold}
Calandra, R.; Peters, J.; Rasmussen, C.~E.; and Deisenroth, M.~P. 2016.
\newblock Manifold Gaussian processes for regression.
\newblock In \emph{2016 International Joint Conference on Neural Networks},
  3338--3345. IEEE.

\bibitem[{Dusenberry et~al.(2020{\natexlab{a}})Dusenberry, Jerfel, Wen, Ma,
  Snoek, Heller, Lakshminarayanan, and Tran}]{dusenberry2020efficient}
Dusenberry, M.; Jerfel, G.; Wen, Y.; Ma, Y.; Snoek, J.; Heller, K.;
  Lakshminarayanan, B.; and Tran, D. 2020{\natexlab{a}}.
\newblock Efficient and scalable bayesian neural nets with rank-1 factors.
\newblock In \emph{International conference on machine learning}, 2782--2792.

\bibitem[{Dusenberry et~al.(2020{\natexlab{b}})Dusenberry, Tran, Choi, Kemp,
  Nixon, Jerfel, Heller, and Dai}]{dusenberry2020analyzing}
Dusenberry, M.~W.; Tran, D.; Choi, E.; Kemp, J.; Nixon, J.; Jerfel, G.; Heller,
  K.; and Dai, A.~M. 2020{\natexlab{b}}.
\newblock Analyzing the role of model uncertainty for electronic health
  records.
\newblock In \emph{ACM Conference on Health, Inference, and Learning},
  204--213.

\bibitem[{Gal and Ghahramani(2016)}]{gal_dropout_2016}
Gal, Y.; and Ghahramani, Z. 2016.
\newblock Dropout as a Bayesian Approximation: Representing Model Uncertainty
  in Deep Learning.
\newblock In \emph{International Conference on Machine Learning}, volume~48 of
  \emph{{JMLR} Workshop and Conference Proceedings}, 1050--1059.

\bibitem[{Gneiting and Raftery(2007)}]{gneiting2007strictly}
Gneiting, T.; and Raftery, A.~E. 2007.
\newblock Strictly proper scoring rules, prediction, and estimation.
\newblock \emph{Journal of the American statistical Association}, 102(477):
  359--378.

\bibitem[{Goel and Chen(2021)}]{goel2021robustness}
Goel, P.; and Chen, L. 2021.
\newblock On the Robustness of Monte Carlo Dropout Trained with Noisy Labels.
\newblock In \emph{IEEE/CVF Conference on Computer Vision and Pattern
  Recognition}, 2219--2228.

\bibitem[{Goodfellow, Shlens, and Szegedy(2015)}]{goodfellow_2015}
Goodfellow, I.; Shlens, J.; and Szegedy, C. 2015.
\newblock Explaining and Harnessing Adversarial Examples.
\newblock In \emph{International Conference on Learning Representations}.

\bibitem[{Handa et~al.(2014)Handa, Whelan, McDonald, and
  Davison}]{handa_etal_2014}
Handa, A.; Whelan, T.; McDonald, J.; and Davison, A.~J. 2014.
\newblock A benchmark for {RGB-D} visual odometry, 3D reconstruction and
  {SLAM}.
\newblock In \emph{International Conference on Robotics and Automation},
  1524--1531. {IEEE}.

\bibitem[{Huang et~al.(2018)Huang, Cheng, Geng, Cao, Zhou, Wang, Lin, and
  Yang}]{huang_2018}
Huang, X.; Cheng, X.; Geng, Q.; Cao, B.; Zhou, D.; Wang, P.; Lin, Y.; and Yang,
  R. 2018.
\newblock The ApolloScape Dataset for Autonomous Driving.
\newblock In \emph{2018 IEEE/CVF Conference on Computer Vision and Pattern
  Recognition Workshops}, 1067--10676.

\bibitem[{Huber(2004)}]{huber2004robust}
Huber, P.~J. 2004.
\newblock \emph{Robust statistics}, volume 523.
\newblock John Wiley \& Sons.

\bibitem[{Jha et~al.(2018)Jha, Raman, Sadigh, and Seshia}]{jha_safe_2018}
Jha, S.; Raman, V.; Sadigh, D.; and Seshia, S.~A. 2018.
\newblock Safe {Autonomy} {Under} {Perception} {Uncertainty} {Using}
  {Chance}-{Constrained} {Temporal} {Logic}.
\newblock \emph{Journal of Automated Reasoning}, 60(1): 43--62.

\bibitem[{Kendall and Gal(2017)}]{kendall_what_2017}
Kendall, A.; and Gal, Y. 2017.
\newblock What {Uncertainties} {Do} {We} {Need} in {Bayesian} {Deep} {Learning}
  for {Computer} {Vision}?
\newblock In \emph{Advances in {Neural} {Information} {Processing} {Systems}},
  volume~30.

\bibitem[{Lakshminarayanan, Pritzel, and
  Blundell(2017)}]{lakshminarayanan_simple_2016}
Lakshminarayanan, B.; Pritzel, A.; and Blundell, C. 2017.
\newblock Simple and Scalable Predictive Uncertainty Estimation using Deep
  Ensembles.
\newblock In \emph{Advances in Neural Information Processing Systems 30, Long
  Beach, CA, {USA}}, 6402--6413.

\bibitem[{Lange, Little, and Taylor(1989)}]{Lange1989RobustSM}
Lange, K.~L.; Little, R. J.~A.; and Taylor, J.~M. 1989.
\newblock Robust Statistical Modeling Using the t Distribution.
\newblock \emph{Journal of the American Statistical Association}, 84: 881--896.

\bibitem[{Lathuili{\`e}re et~al.(2018)Lathuili{\`e}re, Mesejo, Alameda-Pineda,
  and Horaud}]{lathuiliere2018deepgum}
Lathuili{\`e}re, S.; Mesejo, P.; Alameda-Pineda, X.; and Horaud, R. 2018.
\newblock Deepgum: Learning deep robust regression with a gaussian-uniform
  mixture model.
\newblock In \emph{European Conference on Computer Vision}, 202--217.

\bibitem[{Li, Soltanolkotabi, and Oymak(2020)}]{li2020gradient}
Li, M.; Soltanolkotabi, M.; and Oymak, S. 2020.
\newblock Gradient descent with early stopping is provably robust to label
  noise for overparameterized neural networks.
\newblock In \emph{International conference on artificial intelligence and
  statistics}, 4313--4324.

\bibitem[{Liu et~al.(2020)Liu, Lin, Padhy, Tran, Bedrax{-}Weiss, and
  Lakshminarayanan}]{LiuLPTBL20}
Liu, J.~Z.; Lin, Z.; Padhy, S.; Tran, D.; Bedrax{-}Weiss, T.; and
  Lakshminarayanan, B. 2020.
\newblock Simple and Principled Uncertainty Estimation with Deterministic Deep
  Learning via Distance Awareness.
\newblock In \emph{Advances in Neural Information Processing Systems 33:}.

\bibitem[{Malinin and Gales(2018)}]{malinin2018predictive}
Malinin, A.; and Gales, M. 2018.
\newblock Predictive uncertainty estimation via prior networks.
\newblock In \emph{International Conference on Neural Information Processing
  Systems}, 7047--7058.

\bibitem[{Malinin and Gales(2019)}]{malinin_regression_2020}
Malinin, A.; and Gales, M. J.~F. 2019.
\newblock Reverse KL-Divergence Training of Prior Networks: Improved
  Uncertainty and Adversarial Robustness.
\newblock In \emph{Advances in Neural Information Processing Systems}.

\bibitem[{Neverova, Novotn{\'{y}}, and Vedaldi(2019)}]{NeverovaNV19}
Neverova, N.; Novotn{\'{y}}, D.; and Vedaldi, A. 2019.
\newblock Correlated Uncertainty for Learning Dense Correspondences from Noisy
  Labels.
\newblock In \emph{Advances in Neural Information Processing Systems 32},
  918--926.

\bibitem[{Nix and Weigend(1994)}]{nix1994estimating}
Nix, D.~A.; and Weigend, A.~S. 1994.
\newblock Estimating the mean and variance of the target probability
  distribution.
\newblock In \emph{International Conference on Neural Networks}, volume~1,
  55--60. IEEE.

\bibitem[{Northcutt, Jiang, and Chuang(2021)}]{northcutt2021confident}
Northcutt, C.; Jiang, L.; and Chuang, I. 2021.
\newblock Confident learning: Estimating uncertainty in dataset labels.
\newblock \emph{Journal of Artificial Intelligence Research}, 70: 1373--1411.

\bibitem[{Pasi, Jarno, and Aki(2011)}]{JylnkiPasi2011RobustGP}
Pasi, J.; Jarno, V.; and Aki, V. 2011.
\newblock Robust Gaussian Process Regression with a Student-t Likelihood.
\newblock \emph{Journal of Machine Learning Research}.

\bibitem[{Riquelme, Tucker, and Snoek(2018)}]{Riquelme2018Deep}
Riquelme, C.; Tucker, G.; and Snoek, J. 2018.
\newblock Deep Bayesian Bandits Showdown: An Empirical Comparison of Bayesian
  Deep Networks for Thompson Sampling.
\newblock In \emph{6th International Conference on Learning Representations}.

\bibitem[{Ronneberger, Fischer, and Brox(2015)}]{ronneberger2015u}
Ronneberger, O.; Fischer, P.; and Brox, T. 2015.
\newblock U-net: Convolutional networks for biomedical image segmentation.
\newblock In \emph{International Conference on Medical image computing and
  computer-assisted intervention}, 234--241. Springer.

\bibitem[{Schwalbe and Schels(2020)}]{schwalbe_survey_2020}
Schwalbe, G.; and Schels, M. 2020.
\newblock A {Survey} on {Methods} for the {Safety} {Assurance} of {Machine}
  {Learning} {Based} {Systems}.
\newblock In \emph{10th {European} {Congress} on {Embedded} {Real} {Time}
  {Software} and {Systems}}.

\bibitem[{Sensoy, Kaplan, and Kandemir(2018)}]{sensoy_evidential_2018}
Sensoy, M.; Kaplan, L.~M.; and Kandemir, M. 2018.
\newblock Evidential Deep Learning to Quantify Classification Uncertainty.
\newblock In \emph{Advances in Neural Information Processing Systems 31},
  3183--3193.

\bibitem[{Serban, Poll, and Visser(2020)}]{serban_towards_2020}
Serban, A.; Poll, E.; and Visser, J. 2020.
\newblock Towards Using Probabilistic Models to Design Software Systems with
  Inherent Uncertainty.
\newblock In \emph{Software Architecture - 14th European Conference}, volume
  12292 of \emph{Lecture Notes in Computer Science}, 89--97. Springer.

\bibitem[{Silberman et~al.(2012)Silberman, Hoiem, Kohli, and
  Fergus}]{hutchison_indoor_2012}
Silberman, N.; Hoiem, D.; Kohli, P.; and Fergus, R. 2012.
\newblock Indoor Segmentation and Support Inference from RGBD Images.
\newblock In \emph{European Conference on Computer Vision}.

\bibitem[{Snoek et~al.(2015)Snoek, Rippel, Swersky, Kiros, Satish, Sundaram,
  Patwary, Prabhat, and Adams}]{SnoekRSKSSPPA15}
Snoek, J.; Rippel, O.; Swersky, K.; Kiros, R.; Satish, N.; Sundaram, N.;
  Patwary, M. M.~A.; Prabhat; and Adams, R.~P. 2015.
\newblock Scalable Bayesian Optimization Using Deep Neural Networks.
\newblock In \emph{International Conference on Machine Learning}, volume~37 of
  \emph{{JMLR} Workshop and Conference Proceedings}, 2171--2180.

\bibitem[{Tagasovska and Lopez-Paz(2019)}]{tagasovska2019single}
Tagasovska, N.; and Lopez-Paz, D. 2019.
\newblock Single-Model Uncertainties for Deep Learning.
\newblock \emph{Advances in Neural Information Processing Systems}, 32:
  6417--6428.

\bibitem[{Tak, Ellis, and Ghosh(2019)}]{Tak2019RobustAA}
Tak, H.; Ellis, J.~A.; and Ghosh, S.~K. 2019.
\newblock Robust and Accurate Inference via a Mixture of Gaussian and
  Student’s t Errors.
\newblock \emph{Journal of Computational and Graphical Statistics}, 28: 415 --
  426.

\bibitem[{Van~Amersfoort et~al.(2020)Van~Amersfoort, Smith, Teh, and
  Gal}]{van2020uncertainty}
Van~Amersfoort, J.; Smith, L.; Teh, Y.~W.; and Gal, Y. 2020.
\newblock Uncertainty estimation using a single deep deterministic neural
  network.
\newblock In \emph{International Conference on Machine Learning}, 9690--9700.

\bibitem[{Wen, Tran, and Ba(2020)}]{wen2020batchensemble}
Wen, Y.; Tran, D.; and Ba, J. 2020.
\newblock Batchensemble: an alternative approach to efficient ensemble and
  lifelong learning.
\newblock In \emph{International Conference on Learning Representations}.

\bibitem[{West(1984)}]{west1984outlier}
West, M. 1984.
\newblock Outlier models and prior distributions in Bayesian linear regression.
\newblock \emph{Journal of the Royal Statistical Society: Series B
  (Methodological)}, 46(3): 431--439.

\bibitem[{Xia et~al.(2019)Xia, Liu, Wang, Han, Gong, Niu, and
  Sugiyama}]{xia2019anchor}
Xia, X.; Liu, T.; Wang, N.; Han, B.; Gong, C.; Niu, G.; and Sugiyama, M. 2019.
\newblock Are anchor points really indispensable in label-noise learning?
\newblock \emph{Advances in Neural Information Processing Systems}, 32:
  6838--6849.

\end{thebibliography}
\end{document}